\documentclass{article}

\usepackage[preprint]{neurips_2025}

\usepackage[utf8]{inputenc} % allow utf-8 input
\usepackage[T1]{fontenc}    % use 8-bit T1 fonts
\usepackage{hyperref}       % hyperlinks
\usepackage{url}            % simple URL typesetting
\usepackage{booktabs}       % professional-quality tables
\usepackage{amsfonts}       % blackboard math symbols
\usepackage{nicefrac}       % compact symbols for 1/2, etc.
\usepackage{microtype}      % microtypography
\usepackage{xcolor}         % colors
\usepackage{graphicx}
\usepackage{multirow}
\usepackage{float}
\usepackage{adjustbox} % 放在导言区，只需一次
\definecolor{deepblue}{HTML}{27a2c3}

\title{WorldEval: World Model as Real-World Robot Policies Evaluator}

\author{
Yaxuan Li$^{1,2,*}$\thanks{* Equal Contribution. Work done during Yaxuan Li's internship at Midea Group.}     \ \ \ \
Yichen Zhu$^{1,*,\dagger}$\thanks{$\dagger$ Corresponding Author.}  \ \  \ 
Junjie Wen$^{1}$ \ \  
Chaomin Shen$^{2}$ \ \ 
Yi Xu$^{1}$ \ \
\\
\textsuperscript{1} Midea Group
\textsuperscript{2} East China Normal University
\vspace{0.1in}
\\
\href{https://worldeval.github.io}{\color{deepblue}\textbf{worldeval.github.io}} \vspace{-0.3in}
}

\begin{document}

\makeatletter
\let\@oldmaketitle\@maketitle%
\renewcommand{\@maketitle}{\@oldmaketitle
    % \begin{center}
    \begin{figure}[H] % 强制非浮动
        %\captionsetup{type=figure}
        \centering
    \includegraphics[width=1.0\textwidth]{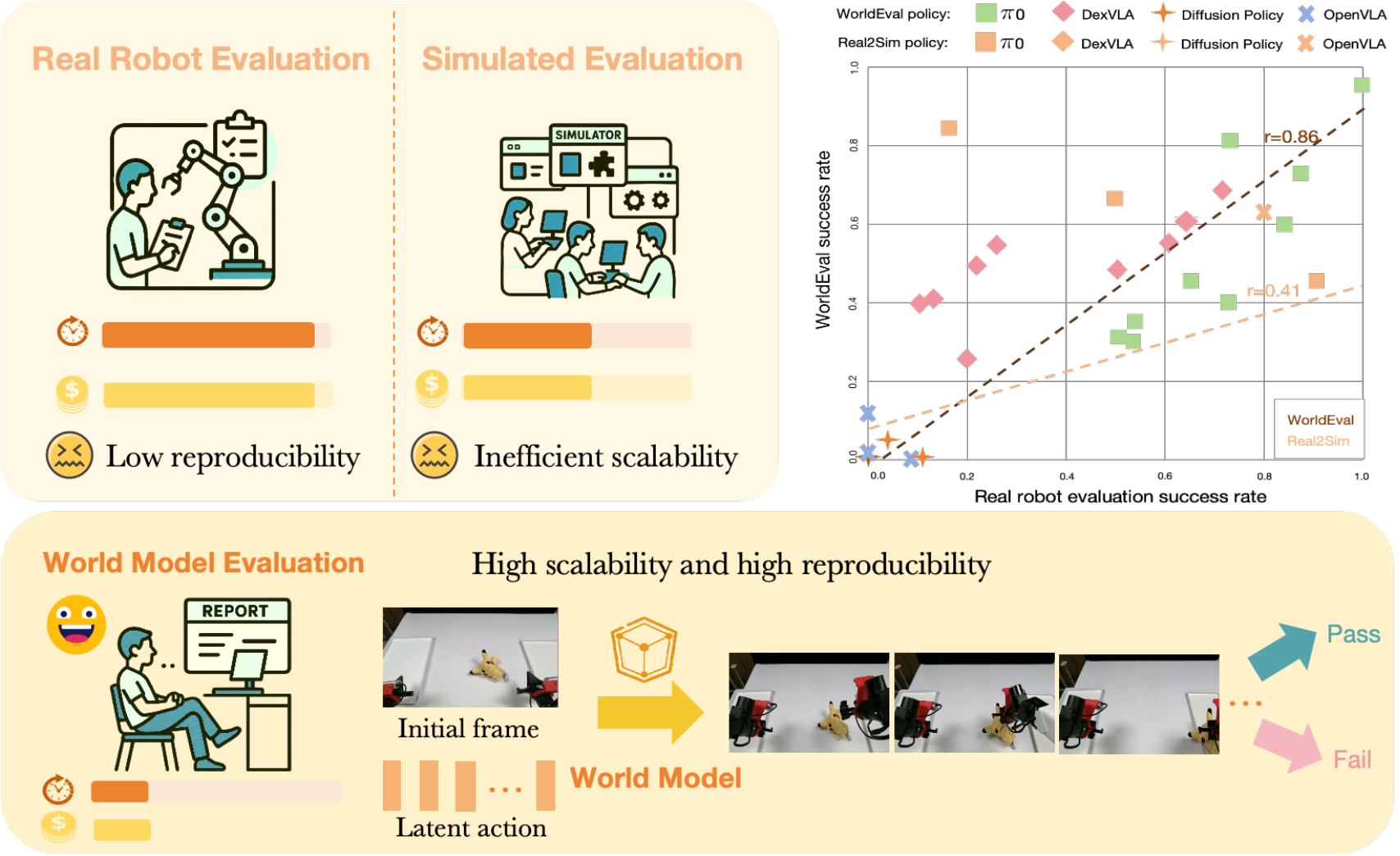}
    \caption{\textbf{WorldEval} is an adaptive, flexible, and reliable approach for evaluating real-world policies across diverse tasks. It demonstrates a strong correlation between its success rates in a world simulator and those observed with real robots.}
    \label{fig:overview}
    % \end{center}
    \end{figure}
    %\vspace{-1em}
}
\def\thanks#1{\protected@xdef\@thanks{\@thanks
        \protect\footnotetext{#1}}}
\makeatother
\maketitle

\begin{abstract}
The field of robotics has made significant strides toward developing generalist robot manipulation policies. However, evaluating these policies in real-world scenarios remains time-consuming and challenging, particularly as the number of tasks scales and environmental conditions change. In this work, we demonstrate that world models can serve as a scalable, reproducible, and reliable proxy for real-world robot policy evaluation. A key challenge is generating accurate policy videos from world models that faithfully reflect the robot’s actions. We observe that directly inputting robot actions or using high-dimensional encoding methods often fails to generate action-following videos. To address this, we propose Policy2Vec, a simple yet effective approach to turn a video generation model into a world simulator that follows latent action to generate the robot video. We then introduce WorldEval, an automated pipeline designed to evaluate real-world robot policies entirely online. WorldEval effectively ranks various robot policies and individual checkpoints within a single policy, and functions as a safety detector to prevent dangerous actions by newly developed robot models. Through comprehensive paired evaluations of manipulation policies in real-world environments, we demonstrate a strong correlation between policy performance in WorldEval and real-world scenarios. Furthermore, our method significantly outperforms popular methods such as real-to-sim approach.
\end{abstract}

\section{Introduction}
The field of robotics has made significant advances towards generalist robot manipulation policies. However, real-world evaluation of such policies faces significant challenges. For instance, training a robot to perform a large number of distinct tasks, such as a thousand unique activities, evaluation on every task would be impractical. It would also be exceptionally difficult to assess whether a new model offers comprehensive improvements over its predecessor, i.e., prior version or early checkpoint. Furthermore, experimentation with novel learning methodologies may inadvertently lead to policies exhibiting incorrect or hazardous behaviors. Testing these behaviors directly in the real world poses significant risks, including potential harm to operators, damage to surrounding equipment, and harm to the robot itself. Consequently, the robotics community urgently requires safe, reliable, and scalable methodologies for evaluating robotic manipulation policies in a controlled environment before real-world deployment.

In this work, we propose the concept of a \textit{world model} as a potential solution, wherein manipulation policies trained on real-world data are evaluated within a world simulator. Such a real-to-video evaluation framework can offer a scalable, reproducible, and insightful complement to traditional real-world assessments. Existing research~\cite{zhang2024pivot,li2025robotic,wu2023daydreamer,huang2025enerverse} on world models~\cite{ha2018world} for robotic manipulation often utilizes video generation as an auxiliary/pre-training objective aimed primarily at improving manipulation performance. In contrast, this research addresses a distinct but equally critical question:

\textit{Can we construct an automated pipeline capable of evaluating and ranking manipulation policies trained on real-world data through a world simulator?}

One viable approach to constructing a world model evaluator involves either directly adopting the action outputs from manipulation policies or employing an explicitly trained action encoder to project robot actions into a discrete, high-dimensional space. This methodology is widely adopted in autonomous driving and video games~\cite{awiszus2021worldgen, guo2025mineworld}. However, in robotic manipulation, the robot's action space can often be semantically ambiguous, such as incremental joint movements, raising an underexplored question about how best to represent these actions. Additionally, interpreting predicted actions from policy models and adequately visualizing them in generated videos poses further challenges. Compounding these issues is the difficulty of accurately capturing the physical properties of objects, especially those previously unseen or situated in novel environments, which significantly hampers scalability.

A central idea in this work is that instead of using the exact action output from the policy model or training a separate network dedicated to action encoding, we propose an alternative approach: the policy network itself inherently provides an effective representation of robot actions. Our insight is that the world model can distinguish between effective and ineffective robotic policies by reflecting their performance directly in the generated video outputs.

To overcome these limitations, we introduce Policy2Vec, a simple strategy that leverages the intrinsic encoding capabilities of the manipulation policy network itself as the action encoder. This innovative approach bypasses the need for external or explicitly trained action encoders by directly harnessing the policy network’s latent representations. Our insight is that policy networks inherently produce informative and discriminative action representations capable of differentiating effective from ineffective policies when visualized through generated videos. Specifically, Policy2Vec injects these policy-generated action vectors into a pre-trained video generation model, effectively converting it into a customized world model tailored to each evaluated policy. Our empirical results demonstrate that these policy-generated encodings robustly reflect policy efficacy through visual outcomes, enabling effective online evaluation of robotic policies.

Building upon Policy2Vec, we introduce WorldEval, an integrated suite of real-to-video evaluation tools specifically designed for robotic manipulation scenarios. Extensive experiments involving multiple publicly available manipulation policies — including Diffusion Policy~\cite{diffusion-policy}, OpenVLA~\cite{openvla}, DexVLA~\cite{wen2025dexvla}, and $\pi_{0}$~\cite{black2024pi_0} — illustrate a strong correlation between policy performance as measured by WorldEval and actual performance observed in real-world experiments. Additionally, we identify the Fréchet Inception Distance (FID) as a lightweight quantitative metric capable of effectively and rapidly ranking different policy models on simple tasks. Furthermore, WorldEval demonstrates potential for assessing robotic policies in entirely new environments and with novel objects, accurately mirroring real-world policy behaviors. Thus, WorldEval represents a significant step towards leveraging world models as reliable and scalable tools for evaluating robotic manipulation policies.

\section{Related Work}

\textbf{Evaluation for robot policies.} Evaluating generalist robotic manipulation policies typically involves extensive testing across diverse tasks and environments. Common evaluation methodologies include real-world and simulation-based approaches~\cite{mees2022calvin, mujoco, mu2021maniskill, gu2023maniskill2, xiang2020sapien, makoviychuk2021isaac, mittal2023orbit, yu2020metaworld, mu2025robotwin, sferrazza2024humanoidbench, grotz2024peract2, liu2023libero,lan2025autobio}. SIMPLER~\cite{li24simpler} introduces a comprehensive approach to evaluate models trained in real-world settings but tested in simulations. AutoEval~\cite{zhou2025autoeval} automates the evaluation process within a controlled laboratory setup using real robots. Nevertheless, these methods have notable limitations. Simulation environments require substantial effort and time to construct, particularly when adapting to different embodiments or scenarios. AutoEval is constrained by its dependence on specific laboratory setups and single robot embodiments. Such evaluation proxies struggle when assessing robot models trained on new embodiments or when evaluating performance across thousands of tasks. The proposed WorldEval framework advances the field by enabling real robot policy evaluations online, regardless of task complexity or robot embodiment type.

\textbf{Robot action representation.} Representation actions is a long-standing substantial problem in robotics research. Broadly, there are
four learning approaches: (i) discretization, where each dimension of the action is divided into discrete bins~\cite{brohan2022rt-1}, 
(ii) function approximation, where actions are parameterized
by a neural network~\cite{diffusion-policy, mobile_aloha,wen2024tinyvla,multimodal_diffusion_transformer,3d_diffuser_actor,wen2024diffusionvla,ze20243d}; (iii) latent space representation,
where actions are encoded in a lower-dimensional space
using unsupervised learning techniques such as variational
autoencoders (VAE)~\cite{wu2024discrete}; and (iv) sequence of language
tokens~\cite{rt-2, openvla, kim2025openvlaoft,pertsch2025fast}, widely used in auto-regressive based vision-language-action models. Our approach explores the function approximation method—specifically, utilizing the policy network itself as an action representation—to enhance the effectiveness of the world model as a robust evaluator.

\textbf{World model for robotics.} The world model predicts future states and outcomes based on learned action representations. There are many application using world model to facilitate the learning of robots on navigation and manipulation~\cite{koh2021pathdreamer, bruce2024genie, bar2024navigation, awiszus2021worldgen,zhu2024maniwm,seo2023multi,ren2023surfer,mendonca2023structured,zhang2024pivot,li2025robotic,wu2023daydreamer,huang2025enerverse,yue2025ewmbench,jiang2025enerverse} Another line of research explores the use of video generation to enhance policy learning\cite{hu2024video,bharadhwaj2024gen2act,chen2024igor,ye2024latent,guo2024prediction,du2023learning,black2023zero,zhao2025cot-vla}. 1X-world~\cite{1xWorldModel} is one of the first work to propose the concept of leverage world model as evaluator for real robot policies. However, they did not present how to achieve this approach. Our method fundamentally differs from the above approach: instead of employing video prediction or a world model to improve robotic policies, our proposed WorldEval aims to train a world simulator capable of performing online evaluation of real robot policies.

\textbf{Imitation learning for robotic control.} Imitation learning methods \cite{diffusion-policy,multimodal_diffusion_transformer,3d_diffuser_actor,ze20243d,ze2023visual,zhu2024scaling,wu2024discrete,li2024cogact,aloha_unleashed,aloha2,mobile_aloha,zhu2024any2policy,Any-point-trajectory,zhu2024retrieval,brohan2022rt-1,rt-affordance,liu2024rdt,jia2024lift3d,wu2024robomind,mail-dp,embodiedcot}, especially vision-language-action models \cite{black2024pi_0,intelligence2025pi_,openvla,rt-2,wen2024tinyvla,wen2024diffusionvla,team2025gemini,ding2025humanoidvla,cui2025openhelix,ding2024quar,liu2025hybridvla,bu2024robodual,bu2025univla,liu2024robomamba,yue2024deer,deng2025graspvla,lin2024datascalinglawsimitation,zhang2024uninavid}, have recently attracted significant attention within the robotics community due to their scalability. Leveraging large-scale datasets combined with powerful models, these approaches efficiently learn within-domain tasks and demonstrate adaptability to numerous out-of-domain scenarios. However, as model capabilities rapidly advance, evaluating real-world policies has become increasingly challenging. Traditional methods of real-world evaluation are not only highly time-consuming and labor-intensive but are also inadequate for assessing performance in novel environments or with unfamiliar objects. To address this limitation, this work proposes utilizing a world simulator for comprehensive and efficient policy evaluation.

\section{WorldEval: World Simulator for Evaluation of Robot Manipulation}
\subsection{Problem Formulation} 
We investigate the application of an action-conditioned world model to evaluate robot manipulation policies and analyze their behavioral characteristics. The aim is not to entirely replace real-world evaluations nor to perfectly replicate real-world policy behaviors via the world model, given the inherent discrepancies between generated video and actual performance conditions. Rather, our objective is to achieve a strong correlation between the relative performance rankings observed in real-world experiments and those derived from the world simulator. Such correlations provide practitioners with an accessible metric to inform iterative improvements in policy design.

In line with this, WorldEval does not seek a precise, one-to-one replication of a policy's real-world behavior. Instead, their purpose is to ensure consistent relative performance between generated video and actual evaluations. Specifically, if a policy outperforms another during real-world testing, this comparative performance should be consistently reflected in WorldEval assessments. Consequently, the world simulator can serve as a reliable proxy, offering actionable feedback for iterative policy refinement. Formally, given two policies, $\pi_{a}$ and $\pi_{b}$, with respective real-world performance measures $R_a$ and $R_b$ (e.g., average success rates across representative task sets), our goal is to develop a world simulator $W$ that generates performance measures $R_{W,a}$ and $R_{W,b}$, strongly correlated with their real-world counterparts.

%We investigate the use of an action-conditioned world model for evaluating robot manipulation policies and analyzing their behavioral modes. Our goal is not to fully replace real-world evaluations or to perfectly replicate real-world behaviors in simulation, given the inherent discrepancies between world models and actual policy performances. Instead, we focus on achieving a strong correlation between the relative performance rankings of policies assessed in the real world and within the world model. This provides practitioners with an accessible and reliable signal to guide policy improvements in their research.

%\section{WorldEval: World Model Evaluated Real-World Robot Policy}
%This section introduce our approaches to designing the pipeline for WorldEval, leverage world model to evaluate real-world robot policy without actually deploy on the real robot. Our method is inspired by rich literature on world model and evaluation framework for robot policies\cite{adamw}. 

\begin{figure}[t]
  \centering
  \includegraphics[width=\linewidth]{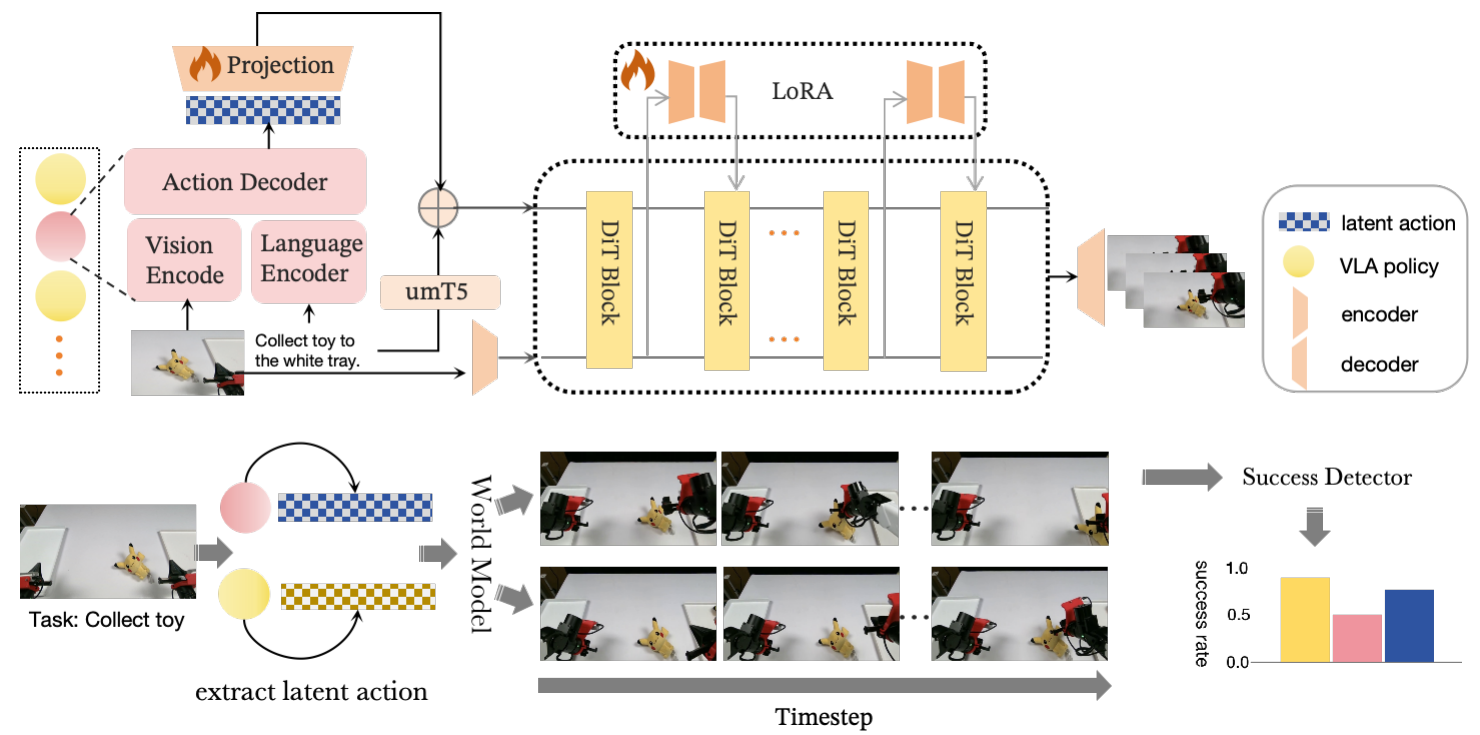}
  \caption{\textbf{Model architecture and evaluation pipeline for WorldEval.} \textbf{Top}: We extract Policy2Vec embeddings from a pool of robot policies and inject them into a pre-trained video generation model, transforming it into a world model. \textbf{Bottom}: The overall WorldEval pipeline for evaluating policy models using the world simulator.}
  \label{fig:structure}
\end{figure}

\subsection{Policy2Vec: Action-Conditioned Robot Video Generation}
Unlike most approaches, which typically focus on generating plausible videos based solely on language instructions and subsequently applying inverse dynamics to derive kinematic information, our goal is to build a world simulator that serves as a policy evaluator. This simulator directly accepts robotic actions as inputs and produces corresponding realistic video outputs. Consequently, this raises the critical question: How can we construct a world model that effectively captures real-world behavior for accurate evaluation?

\textbf{Latent action embedding.} A typical policy model consists of a visual encoder, a language encoder, and an action decoder. Before the action decoder outputs the final action sequence, we can extract latent action embeddings that encode the policy’s internal representations. These latent variables contain rich semantic information in the latent space and can effectively represent the robot’s intended behavior under the current policy.

After obtaining the latent action embedding, we use them as the control signals for a video generation model, along with the language instruction and the initial frame, to guide the generation of future robot video frames. To achieve this, we modified and fine-tuned a pretrained video generation model. In our context, we select WAN 2.1~\cite{wan2025} as our base model due to its strong controllability and its ability to maintain high-quality video synthesis. Other video generation models are also applicable. The WAN 2.1 is built upon the DiT~\cite{peebles2022dit} module, which plays a key role in modeling multimodal conditions to guide the diffusion-based image generation process.

\textbf{Integrating latent action into video generation model.} To align with the WAN 2.1 model, the latent action representations from the policy are first processed through a projection layer to adjust their dimensionality, scaled by a learnable weight $\alpha$, and then added to the language embedding. Note that the policy network itself is not required during fine-tuning; instead, we directly capture latent actions from the policy network along with corresponding initial observations and language instructions, forming a new dataset. The resulting fused control signal is then fed into the DiT module. We performed LoRA~\cite{hu2021lora} fine-tuning on a 14B-scale image-to-video model, primarily training the projection layer and lightly fine-tuning the linear layers within the DiT blocks. This lightweight tuning enables the model to effectively capture the semantics of actions and their representations in the real world. As a result, given the same observation and language instruction, different policies can yield distinct action representations in the latent space. By feeding these representations, along with the first frame and language instruction, into the video generation model, we can intuitively visualize and assess the execution outcomes of each policy, providing a simple and interpretable way to evaluate policy performance. We give a framework overview in Figure~\ref{fig:structure}.

\subsection{Automated Policy Verifier for Policy Success}
Even though our world model generates robot policies via video — making manual evaluation straightforward — it remains crucial to implement an automated process to determine trial success. Specifically, the success classifier is defined as $C_T : S \rightarrow {0, 1}$ and is designed to approximate the ground-truth task success function $T : S \rightarrow {0, 1}$, which maps image states to binary success indicators. For video-based large models, evaluation-based prompts can effectively validate task completion. For example, when provided with a textual prompt such as \textit{"Is this apple place on the plate? Answer yes or no"}, along with the corresponding generated video, the model can accurately classify whether the task was successfully executed. In practice, we utilize Gemini-2.0~\cite{team2023gemini} as our success detector due to its superior video comprehension capabilities.

\section{Experiments}
In this section, we conduct an empirical evaluation of our proposed WorldEval method across a representative set of open-source generalist robotic manipulation policies on multiple real-world robotic tasks. Specifically, our investigation addresses the following research questions:

\begin{itemize}
\item Does the relative performance ranking of different manipulation policies obtained through simulation correlate strongly with their observed performance during real-world evaluations?
\item Does WorldEval offer measurable advantages over simulation-based evaluation?
\item Is WorldEval capable of reliably identifying unsafe policies, thereby preventing catastrophic consequences in real-world deployments?
\item Can WorldEval provide a more cost-effective metric to rapidly evaluate policy models without relying on a success detector?
\item Does Policy2Vec outperform other action encoding methods?
\item Can WorldEval faithfully reflect the actions predicted by the policy network?
\item Beyond comparative performance across diverse policies, can WorldEval accurately replicate internal policy characteristics observed in real-world scenarios, including control frequency and responsiveness to various visual distribution shifts? Furthermore, can WorldEval effectively predict the robustness of policies when subjected to previously unseen distribution shifts in real-world environments?  
\end{itemize}
Due to space constraints, we address the first five questions in the main text, while the remaining questions are discussed in the Appendix, accompanied by additional experiments supporting our claims.

\textbf{Robot setup.} We utilize the bimanual, ALOHA-style robot arm system, AgileX, featuring two arms, each with 6 degrees of freedom (6-DoF). We use a top RealSense 457 camera. This configuration results in a 14-dimensional combined state and action space. Data collection is performed through teleoperation equipment at a frequency of 50 Hz. 

\begin{figure}[t]
  \centering
  \includegraphics[width=0.85\linewidth]{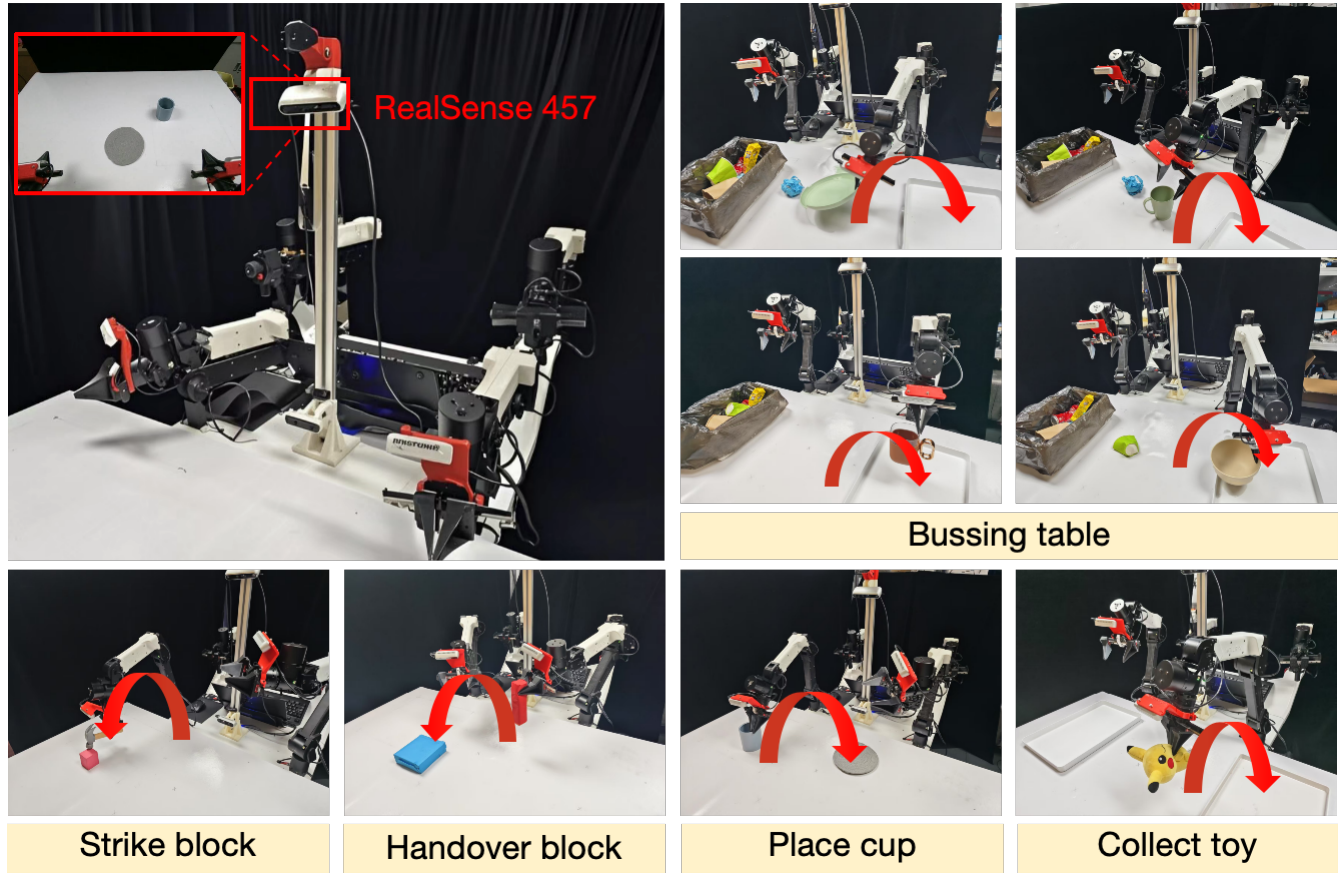}
  \caption{\textbf{Robot and task setup.} We illustrate the robot setup in the top-left figure. We employ a bimanual ALOHA-style robot equipped with top camera views, a RealSense 457. Other cameras are not used in our work.}
  \label{fig:setup}
\end{figure}

\textbf{Task setup.} We designed five experimental tasks: Bussing Table, Collect Toy, Place Cup, Handover Block, and Strike Block. 
\begin{itemize}
    \item \textbf{Bussing Table} is a task that sorts reusable items into a tray and discards trash into a trash bin. This task includes various objects, i.e., a green plate, brown mug, green mug, brown bowl, used paper cup, and blue paper trash, that are placed on the table. This is a challenging task inherited from $\pi_{0}$ and DexVLA. We only use 100 trajectories to finetune the policy network.
    \item \textbf{Collect Toy} is an entirely new task not seen during fine-tuning or pre-training. In this task, the robot must identify and collect previously unseen stuffed toys and place them into a designated area (either the left or right plate).
    \item \textbf{Place Cup}, \textbf{Handover Block}, and \textbf{Strike Block} are three tasks are adapted from the RoboTwin benchmark \cite{mu2025robotwin}. For these three tasks, only 50 trajectories are utilized to fine-tune the policy network. 
\end{itemize}
For each task, we conducted 40 rollouts using the same set of objects. We conduct over 1,000 real-world trials for testing. We present the robot setup and example of tasks in Figure~\ref{fig:setup}.

\begin{table}[t]
  \centering
  \begin{minipage}[t]{0.49\textwidth}
\centering
\caption{\textbf{WorldEval vs. Real-to-Sim evaluation.} WorldEval achieves the lowest MMRV (↓) and highest Pearson $r$ (↑), outperforming real-to-sim.}
\label{tab:real-to-sim_vs_worldeval}
\resizebox{0.99\textwidth}{!}{\begin{tabular}{lcccc}
\toprule
\textbf{Protocol} & \textbf{Place Cup} & \textbf{Strike Block} & \textbf{Handover Block} & \textbf{Avg.} \\
\midrule
\multicolumn{5}{c}{\textbf{MMRV} $\downarrow$} \\
\midrule
Real-to-Sim & 0.332 & 0.266 & 0.187 & 0.261 \\
\textbf{WorldEval} & \textbf{0.000} & \textbf{0.133} & \textbf{0.000} & \textbf{0.044} \\
\midrule
\multicolumn{5}{c}{\textbf{Pearson $r$} $\uparrow$} \\
\midrule
Real-to-Sim & 0.328 & 0.442 & 0.465 & 0.411 \\
\textbf{WorldEval} & \textbf{0.958} & \textbf{0.887} & \textbf{0.980} & \textbf{0.942} \\
\bottomrule
\end{tabular}}
  \end{minipage}
  \begin{minipage}[t]{0.49\textwidth}
    \centering
\caption{\textbf{Policy2Vec versus other encoding methods} We report the Pearson correlation ($r$), MMRV (↓), FID score (↓), and eventual rank across five tasks. Our proposed Policy2Vec achieves the best overall performance.}
\label{tab:encoding_comparison}
\resizebox{0.95\textwidth}{!}{\begin{tabular}{lccccc}
\toprule
\textbf{Encoding Method} & \textbf{Pearson $r$ $\uparrow$} & \textbf{MMRV} $\downarrow$ & \textbf{FID} $\downarrow$ & \textbf{Rank} \\
\midrule
VQVAE~\cite{van2017neural}      & -0.862         &  0.292  & 71.79 & 2 \\
One-hot~\cite{willi2024jafar}     & -0.333         & 0.416 & 75.91 & 3\\
\midrule
\textbf{Policy2Vec (ours)} & \textbf{0.939} & \textbf{0.192} & \textbf{61.33} & \textbf{1} \\
\midrule
\end{tabular}}
  \end{minipage}
  %\vspace{-1em}
\end{table}

\textbf{Implementation details.} We trained the WAN 2.1~\cite{wan2025} model using 1,400 real-world robot trajectories. The training configurations are as follows: initial learning rate of 1e-4, a batch size of 8, a LoRA alpha and rank of 16. The model was trained for 30 epochs. Each trajectory was sampled 3 times, resulting in training samples consisting of 81 frames at $480 \times 480$ resolution.

\textbf{Evaluation metrics.} We use two evaluation metrics to show the correlation between the real-world robot performance and performance on WorldEval. 
\begin{itemize}
    \item The \textbf{Pearson correlation coefficient (Pearson \textit{r})} quantifies the linear relationship between two variables and is widely applied across various domains~\cite{kadian2020sim2real, li24simpler, zhou2025autoeval}. A Pearson correlation coefficient close to 1 signifies a highly effective evaluation proxy, indicating that improvements in real-world success rates directly correspond to linear increases in WorldEval success rates.
    \item \textbf{Mean Maximum Rank Violation (MMRV)} assesses the consistency between real-world and WorldEval policy rankings. For a detailed definition of MMRV, we refer readers to SIMPLER~\cite{li24simpler}. Briefly, MMRV measures the \textit{rank violation} between two policies $\pi_i$ and $\pi_j$, quantifying how severely the proxy evaluation incorrectly ranks policies compared to their actual performance rankings.
\end{itemize}

\subsection{WorldEval Show Strong Performance Correlations with Real-World Evaluations}
\begin{figure}[t]
  \centering
  \includegraphics[width=1\linewidth]{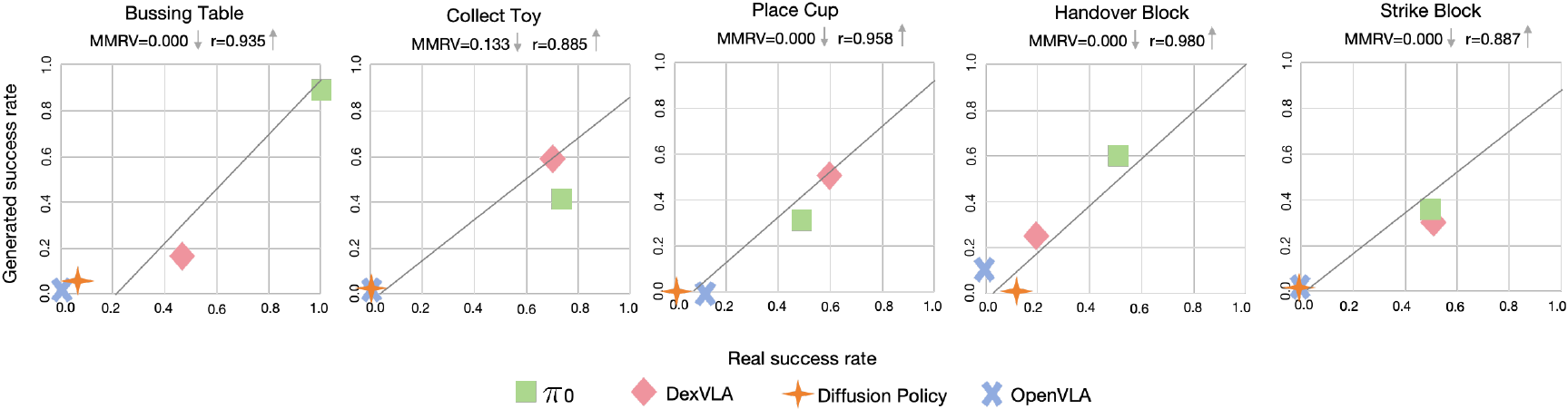}
  \caption{\textbf{Real vs. WorldEval success rates.} WorldEval evaluation setup shows a strong correlation to real policy performance. Good policy evaluation proxies have low MMRV and high Pearson correlation (r).}
  \label{fig:real2world_evaluation}
\end{figure}

In this section, we demonstrate how WorldEval assesses model performance compared to their real-world counterparts. Figure~\ref{fig:real2world_evaluation} summarizes the results of our primary paired evaluations between real-world and generated video scenarios. We observe a robust correlation in relative performance between generated videos and real-world conditions across all four policies evaluated. This correlation supports the effectiveness of WorldEval as a tool for measuring performance during policy development. Specifically, policies exhibiting strong real-world performance, such as DexVLA and $\pi_{0}$, consistently achieve high scores in our generated video evaluations. Conversely, models with lower real-world performance, such as OpenVLA, similarly exhibit lower scores in WorldEval assessments. This consistency is further evidenced by low MMRV metric values and high Pearson r. The experimental results demonstrate that WorldEval serves as a reliable proxy for real-world robot evaluation.

\subsection{Comparison with Real-to-Sim Evaluation}
Evaluating real-world policies in simulation environments is a widely adopted and extensively studied research area, as 
ed in our related work section. In this section, we focus specifically on comparisons with the real-to-sim approach. We utilize RoboTwin~\cite{mu2025robotwin}, which possesses an embodiment identical to our real robot, ensuring consistency across experimental setups. Following the RoboTwin framework, we design and execute three tasks: StrikeBlock, HandoverBlock, and PlaceCup. Policies are trained using real-world data and subsequently evaluated within the RoboTwin simulation environment. Due to limitations regarding equipment and robots specific to the SIMPLER~\cite{li24simpler} method, a direct comparison with SIMPLER was not feasible. However, we apply the real-to-sim techniques proposed by SIMPLER to minimize the visual discrepancies between real-world data and simulation, enabling a fair comparative assessment. 

Table~\ref{tab:real-to-sim_vs_worldeval} presents the experimental results. We observe that, for both metrics—MMRV and Pearson's $r$ — the performance of WorldEval significantly surpasses that of real-to-sim. Despite minimal differences in the robot's physical properties and the extensive efforts to bridge the visual gap between real and simulated environments using real-to-sim techniques from SIMPLER, the simulation still falls short in evaluating real robot policies, even when tasks and objects are identical. These results underscore the effectiveness of our approach, demonstrating its potential as a superior alternative to the traditional real-to-sim method for evaluating real-world robot policies.

\begin{figure}[t]
  \centering
  \includegraphics[width=\linewidth]{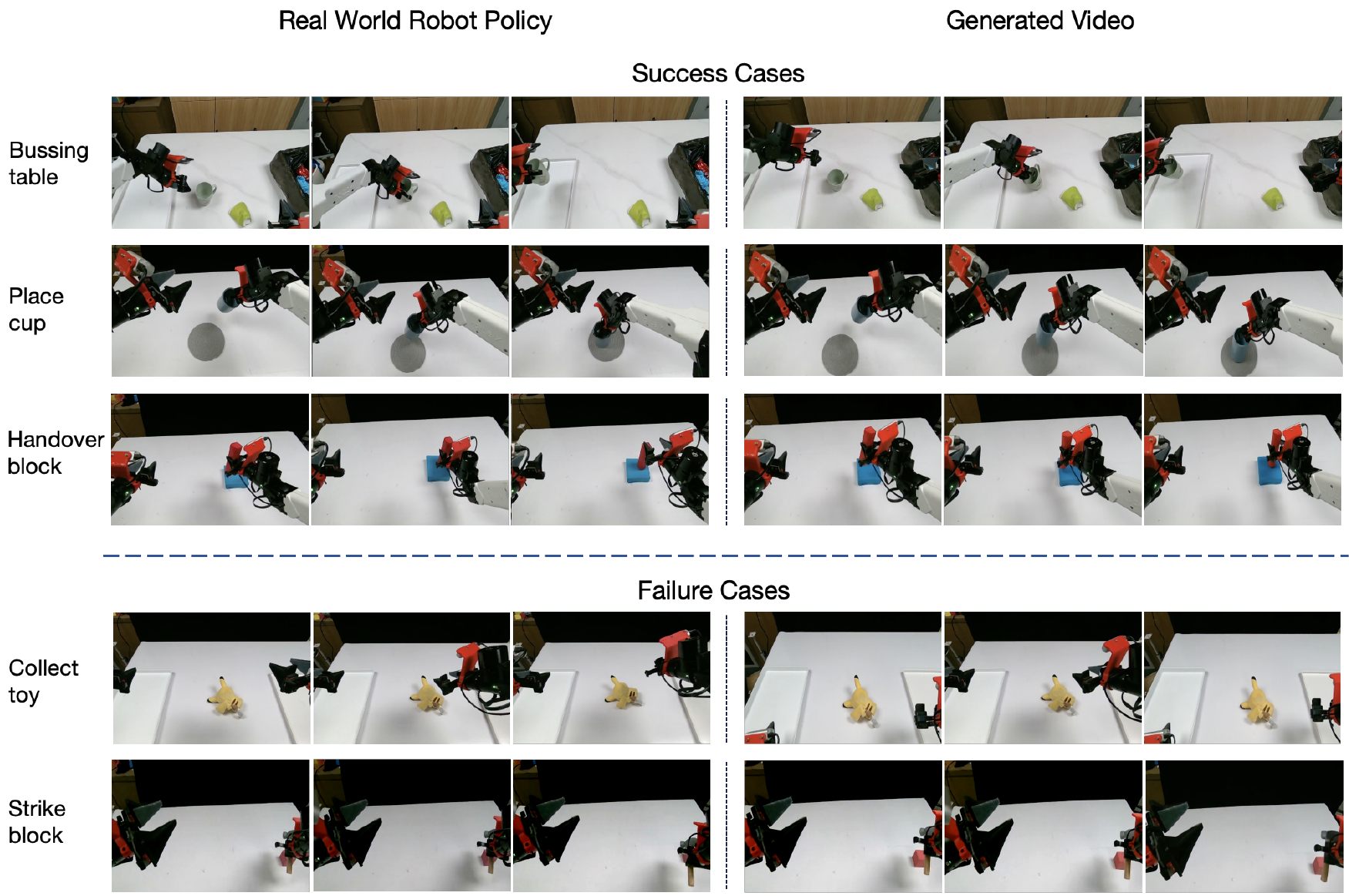}
  \caption{\textbf{Visualization of Real-World Robot Policy and Generated Video Policy.} \textbf{Left:} Real-world robot policy. \textbf{Right:} Generated video policy. \textbf{Top three rows:} Tasks successfully completed by the robot. \textbf{Bottom two rows:} Tasks where the model failed. All policies utilized $\pi_{0}$. The video demo is presented in the supplementary material.}
  \label{fig:visualization}
\end{figure}

\begin{figure}[t]
  \centering
  \includegraphics[width=\linewidth]{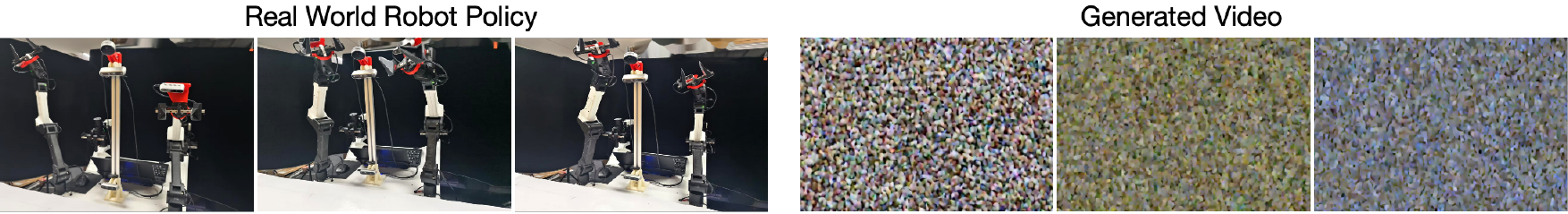}
  \caption{\textbf{WorldEval can serve as a safety detector for robot policies.} On the left is the real-world evaluation, where the robot unexpectedly raises both arms. On the right is the generated video, shown as a mosaic. We show that WorldEval is unable to produce meaningful videos for unsafe robot actions.}
  \label{fig:danger_example}
\end{figure}

\begin{figure}[t]
  \centering
  \includegraphics[width=1\linewidth]{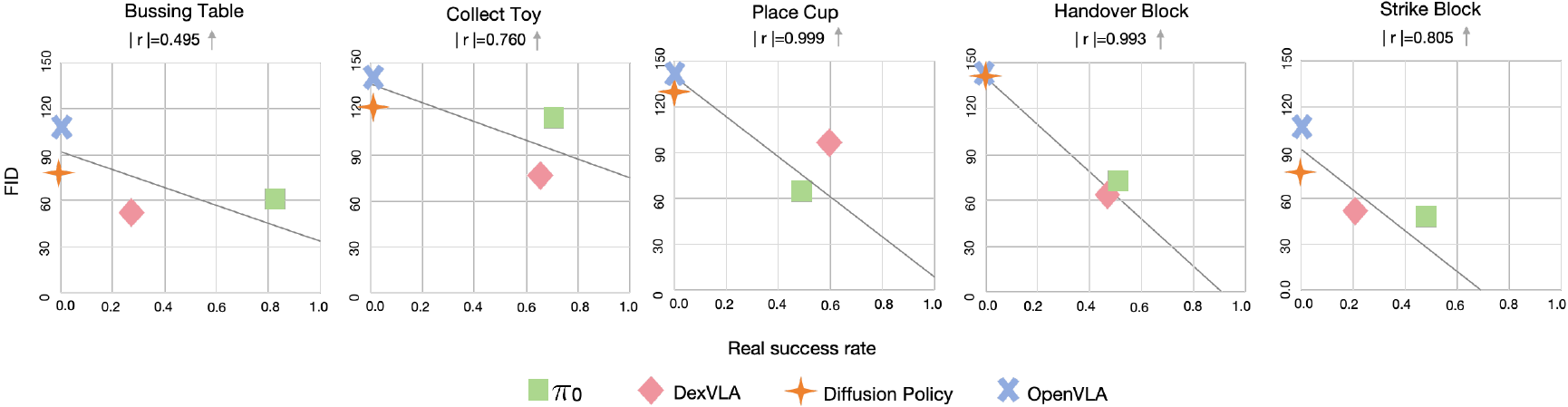}
  \caption{\textbf{Real success rate vs. FID.} We observe that FID serves as a simple and efficient metric for ranking models on basic tasks, making it suitable as a lightweight module for preliminary policy evaluation.}
  \label{fig:real_success_fid_corr}
\end{figure}

\subsection{More Experiments and Ablation Study}
\textbf{Comparison with different action embedding methods.} Policy2Vec serves as the cornerstone of the WorldEval framework, powering its automated evaluation pipeline. To rigorously assess its effectiveness, we conduct a comparative analysis against two alternative action encoding paradigms. First, we implement VQVAE~\cite{van2017neural} — a widely adopted method to discretize continuous actions by vector quantization — following the training protocol established in Discrete Policy~\cite{wu2024discrete}. Second, we evaluate a one-hot encoding baseline, which applies threshold-based categorization to convert continuous actions into discrete representations. As demonstrated in Table~\ref{tab:encoding_comparison}, Policy2Vec achieves superior performance across all metrics, notably reducing the MMRV by 0.1 and 2.24 compared to VQVAE and one-hot encoding, respectively. These findings conclusively demonstrate Policy2Vec’s effectiveness relative to alternative action encoding approaches.

\textbf{Can WorldEval faithfully reflect the actions predicted by the policy network?} We visualize the generated robot videos in Figure~\ref{fig:visualization}, showcasing five tasks. The top three rows illustrate tasks successfully completed by the policy $\pi_{0}$, while the bottom two rows represent tasks where the policy failed during testing. We observe that the actions in the generated videos closely align with those of the real-world robot policies.

Furthermore, we emphasize the "Collect Toy" scenario as an exemplar of our approach's generalization capabilities. This task represents a completely novel challenge, featuring an unseen language instruction, a previously unencountered object (Pikachu), and an entirely new task configuration absent from the training dataset. Despite the inherent novelty and complexity introduced by these factors, the generated video remarkably mirrors the real robot's behavior, successfully executing actions consistent with the desired outcomes. This observation strongly underscores our method's adaptability and robustness, showcasing its scalability to novel, diverse scenes, unfamiliar objects, and previously unseen tasks. 

\textbf{WorldEval as a policy safety detector.} It is common for teams to experiment with new policy architectures, collected data, or training methodologies. However, due to the inherent uncertainty in the training process or model optimization, it is common for the newly learned policy network to produce irregular actions. We demonstrate that WorldEval effectively identifies and discriminates collapsed actions. In Figure~\ref{fig:danger_example} (left), we show a real-world example where the DexVLA model produces a hazardous action, raising both arms straight upwards. In Figure~\ref{fig:danger_example} (right), the corresponding generated video collapses into a mosaic pattern. We hypothesize that the action feature representation for these particular weights is poorly learned, causing WorldEval to fail at extracting meaningful information, leading to the collapsed video output. This observation indicates that WorldEval can effectively discriminate dangerous actions, highlighting its importance as a safeguard against catastrophic outcomes in real-world robot deployment.

\textbf{FID as a lightweight proxy for real robot evaluation.} In addition to leveraging Gemini-2.0 for policy success discrimination in WorldEval, we propose the Fréchet Inception Distance (FID) as a cheap metric for comparative policy evaluation. As illustrated in Figure~\ref{fig:real_success_fid_corr}, FID exhibits strong correlation with real-world success rates across most tasks. This relationship is particularly pronounced in simple manipulation scenarios—such as Place Cup, Handover Block, and Strike Block—where environments contain only one or two objects, confirming FID’s utility for evaluating policies in low-complexity settings. However, the metric’s effectiveness diminishes for intricate tasks like Table Busing, where the Pearson correlation coefficient $r$ decreases significantly, highlighting domain-specific limitations. This result indicates that FID can serve as a preliminary method for evaluating the model on simpler tasks, after which automated methods, such as Gemini-2.0, can be employed for assessing performance on more complex tasks.

\section{Conclusion}
The evaluation of generalist robot manipulation policies in real-world settings presents substantial challenges in scalability, safety, and reproducibility. In this work, we demonstrate that world models can serve as a robust, automated alternative to traditional real-world evaluations by synthesizing policy behaviors in a controlled, risk-free environment. Central to our approach is Policy2Vec, a novel method that leverages the latent representations of policy networks themselves to generate action-following videos, bypassing the limitations of explicit action encoding or high-dimensional inputs. By integrating Policy2Vec with WorldEval, our automated evaluation pipeline, we enable scalable and reliable assessment of robotic policies correlates strongly with real-world performance. Our experiments across diverse manipulation policies, validate WorldEval’s ability to rank policies, identify checkpoint improvements, and detect unsafe behaviors without physical deployment. Notably, WorldEval outperforms conventional real-to-sim approaches and demonstrates generalizability to novel environments and objects. This work underscores the viability of video-based world simulators as a critical tool for the robotics community.

%%%%%%%%%%%%%%%%%%%%%%%%%%%%%%%%%%%%%%%%%%%%%%%%%%%%%%%%%%%%

%%%%%%%%%%%%%%%%%%%%%%%%%%%%%%%%%%%%%%%%%%%%%%%%%%%%%%%%%%%%

\bibliographystyle{unsrt} 
\bibliography{main}

\newpage

\appendix

\section{Limitations}
Although WorldEval offers a promising approach to evaluating robotic manipulation policies, several limitations remain. First, the current framework exclusively relies on action outputs from policy models as control signals for video generation. Incorporating additional modalities such as language instructions or intermediate reasoning could more effectively bridge the gap between simulated evaluations and real-world performance. Secondly, persistent artifacts such as object deformation and visual hallucinations still emerge, especially when evaluating low-performing policies. Issues include unnatural object dynamics, unexpected object appearances or disappearances, overexposure, and robotic arm ghosting effects. These artifacts likely result from insufficient data coverage for low-success scenarios, intrinsic limitations of generative models, or inadequately detailed control inputs. Thirdly, our evaluation pipeline's reliance on pre-trained generative models may restrict adaptability in significantly novel scenarios. Achieving robustness to previously unseen tasks and environments demands further research, potentially involving architectural refinements and additional fine-tuning or retraining steps. Finally, in an ideal scenario, the world model should accurately capture the latent actions in the generated video. However, our current method does not fully achieve this. We believe that training on a large-scale dataset would help mitigate this issue.

\section{More Experiments}

\textbf{WorldEval performance across policy checkpoints.} Table~\ref{tab:ckpt_comparison} summarizes the evaluation results of three different policies at various training steps using WorldEval on a fixed task. The findings reveal a consistent improvement in task success rate as training progresses, suggesting the presence of a clear scaling law for policy performance in real-world scenarios.
At the same time, the FID score decreases with more training steps, suggesting that the latent distribution of actions varies across training stages. Since actions serve as control signals in video generation, these differences significantly affect video quality. Specifically, policies that are not sufficiently trained produce latent actions that diverge from those of well-trained policies, thereby degrading the performance of the video reconstruction model. This is especially relevant because we use latent actions from the final policy checkpoint to fine-tune the video generation model.
Moreover, we find that FID exhibits a strong correlation with real-world task success rates. This suggests that FID can serve as an initial metric for evaluating model performance on simple tasks.

\begin{table}[h]
\centering
\footnotesize
\caption{\textbf{Performance comparison across different policies and training steps.} We report FID and average success rate (Succ.) for three tasks in the Bussing table scenario. Task A, B, and C involve placing a \textit{brown bowl}, \textit{green plate}, and \textit{brown mug}, respectively.}
\label{tab:ckpt_comparison}
\begin{tabular}{l c | cc cc cc}
\toprule
\textbf{Method} & \textbf{Step} 
& \multicolumn{2}{c}{\textbf{Task A}} 
& \multicolumn{2}{c}{\textbf{Task B}} 
& \multicolumn{2}{c}{\textbf{Task C}} \\
 & & FID & Succ. & FID & Succ. & FID & Succ. \\
\midrule
\multirow{3}{*}{$\pi_{0}$} 
  & 5k          & 81.62 & 0.725 & 67.57 & 0.075 & 72.52 & 0.300 \\
  & 30k         & 78.07 & 0.775 & 65.30 & 0.150 & 69.51 & 0.450 \\
  & Final (60k) & 74.65 & 0.925 & 63.45 & 0.600 & 65.39 & 0.750 \\
\midrule
\multirow{3}{*}{DexVLA} 
  & 5k          & 488.26 & 0.000 & 489.83 & 0.000 & 530.71 & 0.000 \\
  & 10k         & 101.43 & 0.000 & 99.50 & 0.000 & 99.65 & 0.000 \\
  & Final (60k) & 57.96  & 0.125 & 56.60  & 0.250 & 69.56  & 0.450 \\
\midrule
\multirow{3}{*}{Diffusion Policy} 
  & 5k          & 83.60 & 0.000 & 86.13 & 0.000 & 86.41 & 0.000 \\
  & 30k         & 80.99 & 0.000 & 80.52 & 0.000 & 85.32 & 0.000 \\
  & Final (40k) & 82.96 & 0.000 & 78.96 & 0.000 & 84.87 & 0.000 \\
\bottomrule
\end{tabular}
\end{table}

\begin{figure}[t]
  \centering
  \includegraphics[width=\linewidth]{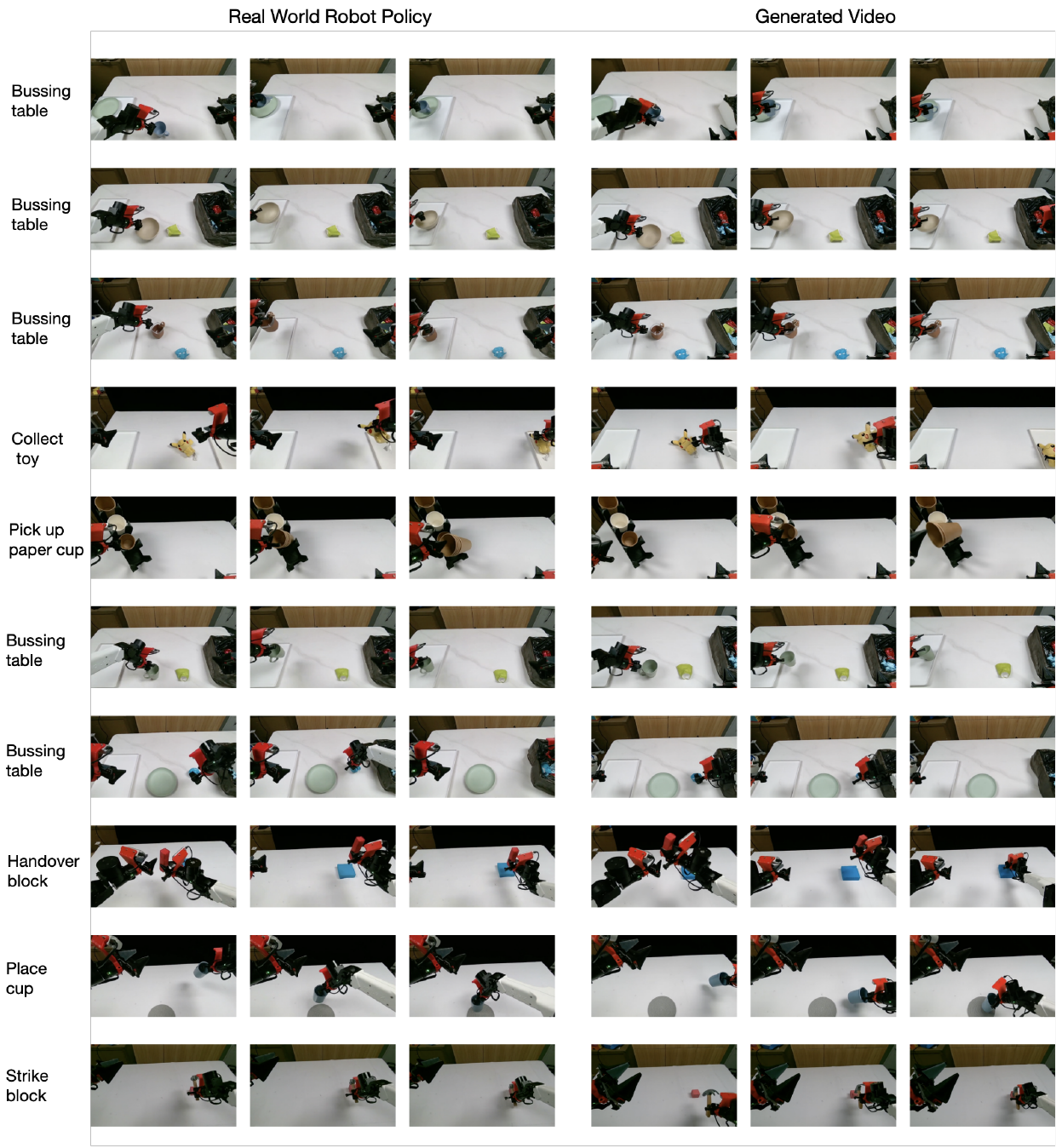}
  \caption{\textbf{Visualization of Real-World Robot Policy and Generated Video Policy.} \textbf{Left:} Real-world robot policy. \textbf{Right:} Generated video policy. Tasks successfully completed by the robot.}
  \label{fig:success_visual}
\end{figure}

\begin{figure}[t]
  \centering
  \includegraphics[width=\linewidth]{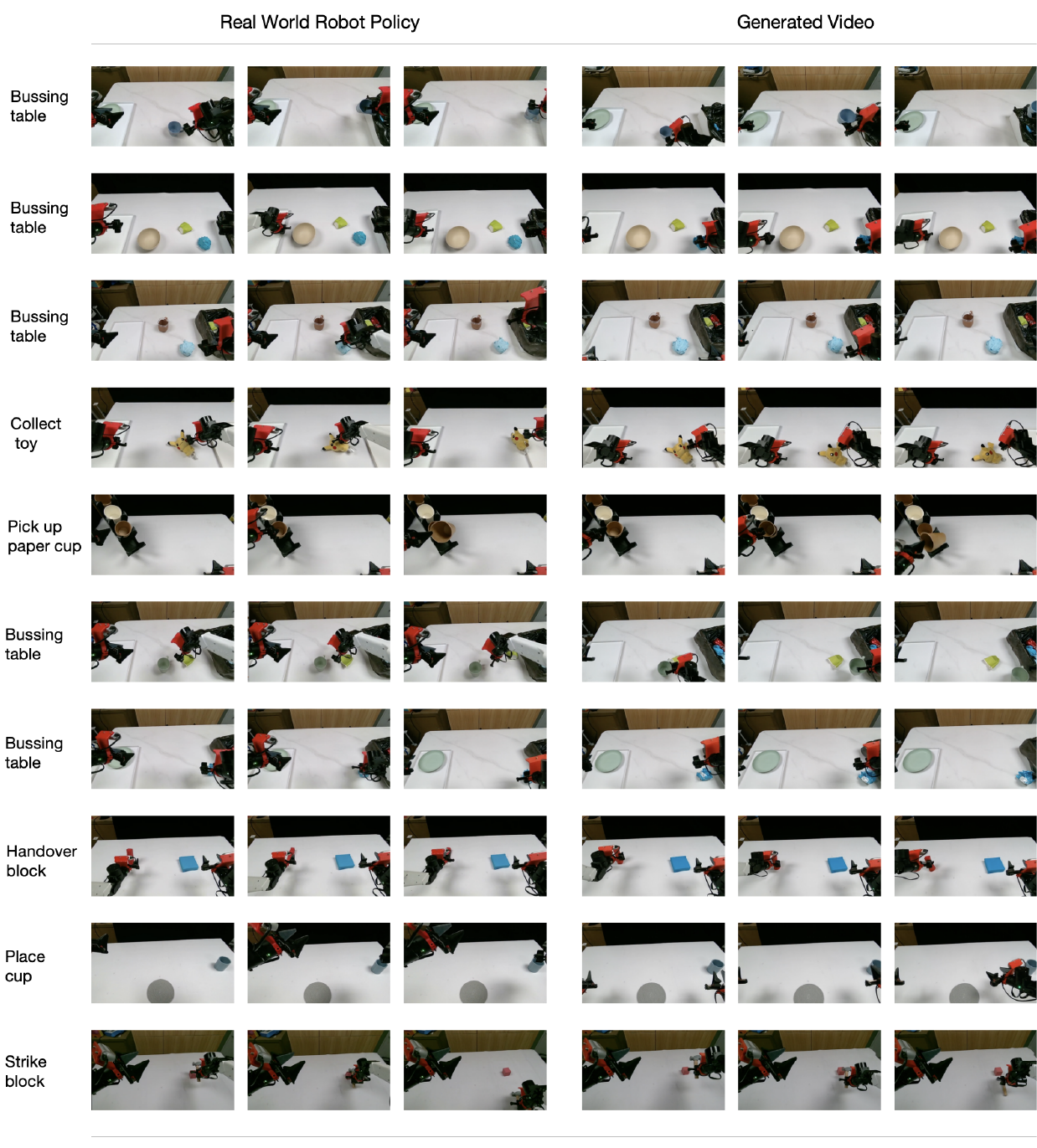}
  \caption{\textbf{Visualization of Real-World Robot Policy and Generated Video Policy.} \textbf{Left:} Real-world robot policy. \textbf{Right:} Generated video policy. Tasks where the model failed. }
  \label{fig:failure_visual}
\end{figure}

\begin{figure}[t]
  \centering
  \includegraphics[width=\linewidth]{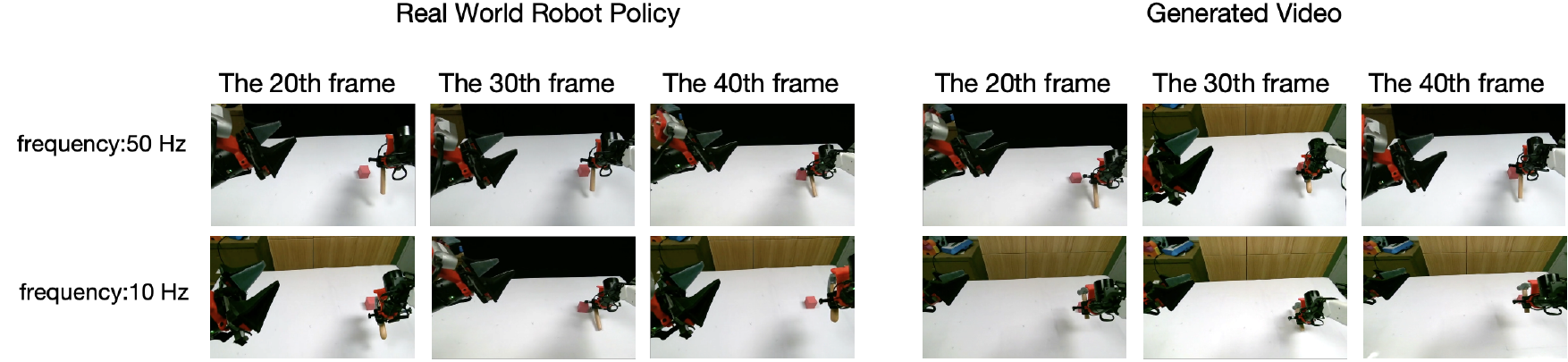}
  \caption{WorldEval-generated policy networks trained on data collected at different frequencies.}
  \label{fig:10hz_vs_50hz_visual}
\end{figure}

\textbf{Effective Prediction of Policy Robustness Under Distribution Shifts.}
We further explore WorldEval by evaluating our policy in unseen environments and with novel objects on new tasks. Specifically, in the main text, the "collect toy" task represents a scenario involving both unseen objects and instructions. Our previous findings demonstrate that WorldEval maintains robust video reconstruction capabilities and exhibits a strong correlation between real-to-video evaluations and real-world testing, even for these entirely new tasks.

Additionally, we extend our evaluation to scenarios with unseen backgrounds. Although the policy network was originally trained using data collected exclusively in a laboratory environment, we assessed its performance in distinctly different settings: an office, a living room, and a kitchen. We conducted tests involving three tasks—placing a cup, striking a block, and handing over a block—in these novel environments. The MMRV for tasks in the office, living room, and kitchen was 0.047, only slightly lower than the in-domain (laboratory) environment performance of 0.044, but significantly higher than typical real-to-simulation approaches. Moreover, the Pearson $r$ remained high at 0.927. These results collectively indicate a robust correlation and consistent performance even when tasks involve novel instructions and manipulating previously unseen objects in new environments.

\section{More on Experimental Setup}

\textbf{Implementation details.} We trained the WAN2.1 14B image-to-video model using 1,400 real-world robot trajectories. The training configuration includes an initial learning rate of 1e-4, a batch size of 8, and a LoRA alpha and rank of 16. The model was trained for 30 epochs. Each trajectory was sampled three times, resulting in training samples consisting of 81 frames at a resolution of 480 $\times$ 480. The training was conducted using 8 H800 GPUs with data parallelism and took approximately 11 hours. Each video is generated using 50 diffusion steps.

\textbf{Evaluation on Robotwin.} To address the domain gap between simulation and the real world, we employ advanced computer vision techniques to bridge the visual disparity. While SIMPLER modifies simulation textures, we take a different approach by utilizing sim-to-real visual models. Additionally, we integrate the MidJourney API, which transforms simulation images into more realistic representations of the real world. This method has proven to be highly effective. Upon receiving the current observation from the simulation benchmark, we call MidJourney to convert the image into a more realistic version before feeding it into the policy network, which then generates the next sequence of actions.

\textbf{More description on tasks.} We designed five experimental tasks: Bussing Table, Collect Toy, Place Cup, Handover Block, and Strike Block.
\begin{itemize}
    \item \textbf{Bussing Table} is a task that requires the robot to sort reusable items into a tray and dispose of trash into a trash bin. The scene includes a variety of objects placed on the table, such as a green plate, brown mug, green mug, brown bowl, used paper cup, and blue paper trash. This is a challenging task derived from $\pi_0$ and DexVLA. The natural language instruction for this task is: \textit{``Clean the table.''}. This task includes 100 trajectories. During evaluation, object positions are randomized on the table to ensure diverse testing conditions.
    \item \textbf{Collect Toy} is a completely novel task that was not encountered during either fine-tuning or pre-training. In this task, the robot must identify and collect previously unseen stuffed toys and place them into a designated area—either the left or right tray. The tabletop scene setup is also out-of-domain for the policy. The language instruction is: \textit{``Collect the toy to the right/left tray.''}
    \item \textbf{Place Cup}, a blue empty cup and a cup mat are placed at random positions on the table in front of the robot. The robot is required to place the cup onto the mat. The language instruction is: \textit{``Place the empty blue cup to the cup mat.''}. This task includes 50 trajectories.
    \item \textbf{Handover Block}, the robot must use its left arm to grasp a red rectangular block and pass it to the right arm, which then places the block onto a blue mat. The language instruction is: \textit{``Pass the red block to the right arm to place it on the blue mat.''}. This task is moderately difficult, requiring coordinated bi-arm manipulation and precise control of the block's orientation to avoid tipping it over during handover and placement. This task includes 50 trajectories.
    \item \textbf{Strike Block}, the robot is instructed to pick up a hammer and strike a red block placed in front of it. The language instruction is: \textit{``Pick up the hammer, then strike the red block.''}. This task tests the policy's ability to localize and interact with small objects in the visual field, as the red block is relatively small. This task includes 50 trajectories.
\end{itemize}

\section{More Visualization}
We give more visualization for the real world image and the generated video in this section. Figure~\ref{fig:success_visual} presents the success cases, and Figure~\ref{fig:failure_visual} presents the failure cases. 

\textbf{WorldEval with different data collection frequency.} We also analyze whether differences in data collection frequency affect the quality of the generated video. In  Figure~\ref{fig:10hz_vs_50hz_visual}, the first row displays data collected at 10Hz, while the second row shows data collected at 50Hz. Two policy networks are trained separately on these datasets, each containing an equal number of trajectories. Note that a single WorldEval model is utilized for evaluation. Results indicate that WorldEval effectively captures variations in data frequency; the model trained on 50Hz data produces smoother actions that are less segmented, resulting in slower and more fluid video generation. This further demonstrates that WorldEval accurately interprets latent actions.

\end{document}